\documentclass[conference]{IEEEtran}
\IEEEoverridecommandlockouts
\usepackage{cite}
\usepackage{amsmath,amssymb,amsfonts}
\usepackage{algorithmic}
\usepackage{graphicx}
\usepackage{textcomp}
\usepackage{xcolor}
\usepackage{multirow}
\def\BibTeX{{\rm B\kern-.05em{\sc i\kern-.025em b}\kern-.08em
    T\kern-.1667em\lower.7ex\hbox{E}\kern-.125emX}}
\DeclareRobustCommand*{\IEEEauthorrefmark}[1]{%
\raisebox{0pt}[0pt][0pt]{\textsuperscript{\footnotesize\ensuremath{#1}}}}
\begin{document}

\title{Semantic Palette-Guided Color Propagation}

\author{\IEEEauthorblockN{Zi-Yu Zhang\IEEEauthorrefmark{1}, Bing-Feng Seng\IEEEauthorrefmark{1,2}\thanks{Zi-Yu Zhang and Bing-Feng Seng contribute equally to this work.}, Ya-Feng Du\IEEEauthorrefmark{1}, Kang Li\IEEEauthorrefmark{1}, Zhe-Cheng Wang\IEEEauthorrefmark{1,2}, Zheng-Jun Du\IEEEauthorrefmark{1,2}\thanks{Zheng-Jun Du  is the corresponding author.}}\IEEEauthorblockA{\IEEEauthorrefmark{1}School of Computer Technology and Application, Qinghai University, Xining, China}
\IEEEauthorblockA{\IEEEauthorrefmark{2}Qinghai Provincial Laboratory for Intelligent Computing and Application, Xining, China}}

\maketitle

\begin{abstract}
Color propagation aims to extend local color edits to similar regions across the input image. Conventional approaches often rely on low-level visual cues such as color, texture, or lightness to measure pixel similarity, making it difficult to achieve content-aware color propagation. While some recent approaches attempt to introduce semantic information into color editing, but often lead to unnatural, global color change in color adjustments. To overcome these limitations, we present a semantic palette-guided approach for color propagation. We first extract a semantic palette from an input image. Then, we solve an edited palette by minimizing a well-designed energy function based on user edits. Finally, local edits are accurately propagated to regions that share similar semantics via the solved palette. Our approach enables efficient yet accurate pixel-level color editing and ensures that local color changes are propagated in a content-aware manner. Extensive experiments demonstrated the effectiveness of our method.

\end{abstract}


\section{Introduction} 
\label{sec:introduction}

In recent years, with the rapid development of social media, hundreds of millions of pictures rapidly spread on the Internet every day. Thus a large number of application scenarios have placed new demands on the ease of use and efficiency of image processing. Color editing is a key component in image processing, and related techniques have been widely used in the fields of graphic design, image enhancement, recoloring, etc. To achieve intuitive and efficient color editing,  researchers have proposed two kinds of methods: edit propagation and palette-based image recoloring.

Edit propagation aims to extend local user edits to other regions of an image that share similar color, texture or brightness. Levin et al.\cite{LAL*04} introduced the first approach of edit propagation  to grayscale image colorization. They formulated the colorization task as a energy optimization problem, such that pixels with similar brightness to user strokes receive similar colors. Following closely behind, edit propagation was successfully extended to tonal adjustment \cite{lischinski06} and material editing \cite{PFL*07}. Later, An et al. \cite{AF08} proposed a unified framework ``Appprop" for alerting appearance of complex spatially-varying datasets, such as images and physically measured materials. However, their method is computationally and storage costly due to measuring the similarity of all pixel pairs. To overcome this limitation, Xu et al. \cite{XLJ*09} presented a K-D tree-based approach to edit propagation, which significantly improves computational efficiency and reduces memory overhead. Besides, Li et al. \cite{LJH10} re-formulated the edit propagation as a function interpolation problem in a high-dimensional space, which can be efficiently solved through radial basis functions. Similarly, various other studies, including those based on clustering \cite{bie2011real}, 
bilateral grids \cite{LOG*19}, and color palette \cite{Xia2024} have been introduced to facilitate real-time color editing. Most recently, Endo et al.\cite{YSY*16} proposed the first deep learning-based network, ``Deepprop", to edit propagation, however, it needs density and fine tuned user stokes, and cannot accurately propagate user edits to nearby regions. Gui et al.\cite{GZ20} introduced a CNN-based method that formulates edit propagation as a  multi-class classification problem, but it is computational expensive and fails to propagate local edits to  distant pixels with similar semantics. 

Palette-based image recoloring is a recent popular method in color editing. It allows user to adjust the appearance of an image by editing a small set of dominant colors. The pioneering palette-based approach was proposed by Chang et al.\cite{CFL*15}. They first apply modified K-means clustering to create a color palette from an image, and then design a radial basis function-based \cite{buhmann2000} algorithm to map palette colors' change to the whole image. Followed by this, Zhang et al.\cite{ZXS*17} adopted an energy optimization method to compute the weights of pixel color to palette colors, producing visually natural and vivid recoloring results. Recently, Zhang et al.\cite{zhang2021blind} proposed a blind color separation model to concurrently solve the color palette and corresponding mixing weights of pixel colors to the palette. Tan et al.\cite{TLG16} proposed the first convex hull based approach for palette extraction and image recoloring. They first project the input image into RGB-space, and then calculate the simplified convex hull and use its vertices as the color palette, at last, each pixel's color is expressed as a weighted sum of the palette colors. During color editing, the user recolors an image by adjusting its palette colors. Next, Tan et al.\cite{TEG18_0} extended this method to RGBXY-space, achieving more efficient recoloring. Wang et al.\cite{WLX19} proposed a geometric approach to improve the representativeness of the palette colors generated by Tan et al.\cite{TEG18_0}. Sun et al.\cite{sun2023building} introduced a coarse-to-fine convex hull method with added auxiliary vertices, to further reduce the reconstruction error of convex hull-based approaches, while making the palette colors better representative. Du et al.\cite{DLX*21} extended the convex hull-based image recoloring to video scenario. They employ the skew polytope in RGBT-space to present the color palette of a video, achieving natural and time-varying video recoloring.
\begin{figure*}[htbp]
\includegraphics[width=1\textwidth]{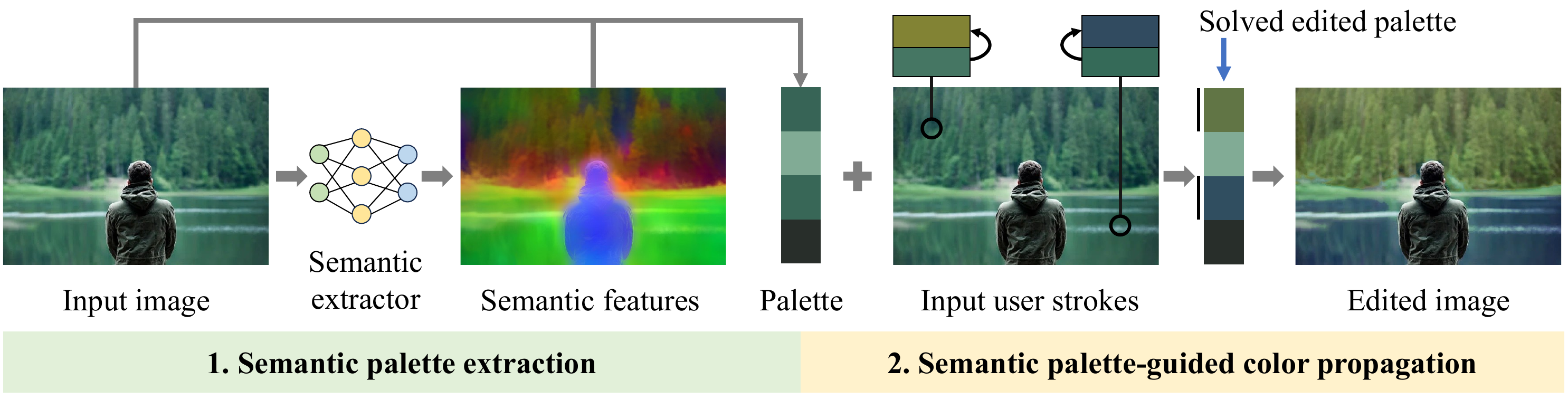}
\caption{Pipeline of our method.}
\label{fig:pipeline}
\vspace{-0.4cm}
\end{figure*}
Most recently, Chao et al.\cite{chao2023colorful} introduced ``ColorfulCurves" which bridges the gap between palette-based and tone curve editing. To enhance local control for palette-based image recoloring, Chao et al.\cite{chao2023loco} proposed ``LoCoPalettes" that incorporates palette-based image editing with image-space constraints and semantic hierarchies. Du et al.\cite{du2024palette} presented a palette-based approach for content-aware image recoloring, but this method neither generates palettes adaptively nor supports pixel-level color editing.


Through literature study, we found that most edit propagation algorithms offer intuitive interactions, but they usually require users to provide dense and fine-tuned strokes. Moreover, these techniques typically rely solely on low-level features such as color, texture, or material properties to measure pixel similarity, making it difficult to achieve content-aware color propagation. In contrast, most palette-based approaches are more efficient, but they do not support pixel-level color editing. Although some recent approaches attempt to incorporate semantic information for color editing, but fail to capture the similarity between pixels that are spatially distant, resulting in unintended global color shifts during color editing. To overcome these limitations, in this paper, we propose a novel semantic palette-guided approach for content-aware color propagation. In summary, compared with existing methods, we made the following contributions:
\begin{itemize}
\item Our method automatically extracts a semantic palette from an input image, such that each palette entry could correspond to one or more semantically similar regions. 
\item With the palette, user strokes can be efficiently and accurately extended to regions with similar semantics.
\item Our method requires only very sparse and non-fine-tuned user strokes for content-aware color propagation.
\end{itemize}



\section{Method} 
\label{sec:method}
Our goal is to propagate sparse, local color edits to other regions that share similar semantics within an image, achieving intuitive, efficient yet content-aware color propagation. To this end, we propose a two-step approach which takes an image and some user strokes as input, outputs an edited image. 

The pipeline of our two-step method is illustrated in Fig.~\ref{fig:pipeline}. The first step is semantic palette extraction. Given an input image, we first employ a neural network to extract per-pixel semantic features. We then project the input image into a high-dimensional feature space that fuses color and semantic information, to extract its semantic palette. The second step is semantic palette-guided color propagation. Given an input image, the extracted palette along with user strokes, we first seek an edited palette aligning with local color change by minimizing a well-designed energy function. We then extend edited palette colors to semantically similar objects or regions, resulting in content-aware color propagation.

\subsection{Semantic Palette Extraction}
\label{sec:semantic-palette-extraction}
This step aims to extract a semantic palette $P$ from an image $I$, where each palette entry $P_i$ corresponds to one or more objects or regions that share similar semantics. To this end, we first employ a neural network to extract pixel-wise semantic features. Followed by this, the input image is projected into a feature space that fuses low-level visual and high-level semantic features. At last, a semantic palette is created in this space via a modified k-means algorithm.

We employ a neural network introduced by Aksoy et al.\cite{YOA*17} to extract per-pixel semantic features. This network takes an image as input and outputs a 128-dimensional semantic feature for each pixel. To reduce computational and storage burdens, as well as to eliminate redundant information, we adopt Principal Component Analysis (PCA) to condense the feature dimensionality of each pixel down to three dimensions. Next, we combine the color and semantic of each pixel to build a feature space. So that any pixel $I_i$ can be viewed as a 6D feature point $I_i = (r_i,g_i,b_i,\kappa_i^1,\kappa_i^2,\kappa_i^3)$. For sake of description, we use $I_i.C = (r_i,g_i,b_i)$ and $I_i.S = (\kappa_i^1,\kappa_i^2,\kappa_i^3)$ to represent the color and semantic components of $I_i$, respectively. 


\begin{figure*}[!t]
\includegraphics[width=1\textwidth]{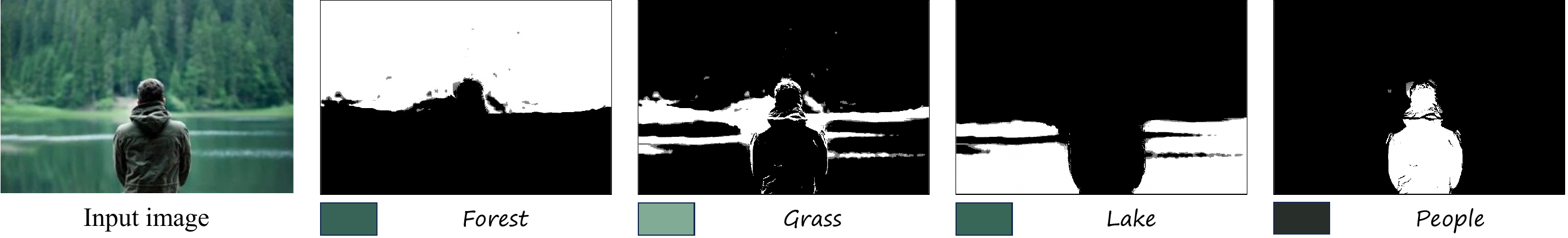}
\caption{Visualization of similarity weights. For this example, we provide the input image, similarity weights of all pixels to each semantic palette entry, and the color component of each palette entry. This visualization illustrates that each palette entry is associated with certain semantically similar regions.
}
\label{fig:weights-vis}
\vspace{-0.4cm}
\end{figure*}

Next, we design a modified k-means algorithm to automatically select a small set of representative feature points to form a semantic palette, from the feature space corresponds to the input image. Firstly, to speed up the clustering process, we employ the Simple Linear Iterative Clustering (SLIC) algorithm\cite{SLIC12} to segment the input image into superpixels, and use those pixels located at the centroids of these superpixels as the sampling points. Secondly, we design a scheme to automatically select the initial centers without specifying the number of classes. Specifically, we begin by assigning each feature point $I_i$ an importance weight $\pi_i$, setting its initial value as the pixel count within the $i$-th superpixel. Next, we select the sampling point with maximum weight and update any other sampling point $I_j$'s weight $\pi_j$ by:
\begin{equation}
    \pi_j = (1 - \exp(-d(I_i, I_j)^2)) \cdot \pi_j.
\end{equation}
Where $d(I_i, I_j)$ denotes the distance between $I_i$ and $I_j$:
\begin{equation}
    d(I_i, I_j) = w_c  \left\|I_i.C - I_j.C \right\|_2 + w_s\left\|I_i.S - I_j.S\right\|_2.
\label{eq:feature-distance}
\end{equation}
Where $w_c$ and $w_s$ are weight parameters used to balance the relative contributions of color and semantic distances. Next, we iteratively select the initial centers until the current maximum weight is less than a predefined threshold $t$. Once the initial centers are determined, we could perform k-means on all sapling feature points, and the resulting convergent cluster centers act as the semantic palette.

Assuming the extracted palette is $P=\{P_1,P_2,\cdots,P_k\}$, now we can compute the similarity weight of any pixel $x$ to the palette entry $P_i$. We measure this similarity using both color and semantic information, so that pixels in the same region that are similar in color but slightly different in semantics can be brought closer by color information. We first define a Radial Basis Function for each palette entry as:
\begin{equation}
\small{
   \phi(x,P_i) = \exp\left(\frac{-\|x.C-P_i.C\|^2_2}{2\sigma_c^2}\right)  \exp\left(\frac{-\|x.S-P_i.S\|^2_2}{2\sigma_s^2}\right).
   }
\end{equation}
Here the standard deviation for color, denoted as 
$\sigma_c$, and that for semantic features, denoted as $\sigma_s$, are both determined by averaging the color and semantic features across all palette entries. Then the similarity of any pixel $x$ to the palette entry $P_i$ can be defined as:
\begin{equation}
\setlength\abovedisplayskip{3pt}
    f_i(x) = \sum_{j=1}^k\lambda_{i,j}\phi(x,P_j).
\setlength\belowdisplayskip{3pt}
\end{equation}
Where $\lambda_{i,j}$ is a coefficient which can be obtained by solving a linear system\cite{CFL*15}. And the normalized similarity weight of any pixel $x$ to the palette entry $P_i$ is given by:
\begin{equation}
    w_i(x) = \frac{f_i(x)}{\sum_{j=1}^k f_j(x)}.
\end{equation}
Based on this, we can calculate the similarity weights of all pixels with respect to each palette entry. The visualization of similarity weights are shown in Fig.\ref{fig:weights-vis}. As can be seen, each palette entry could correspond to some specific regions.

\subsection{Semantic Palette-Guided Color Propagation}
\label{sec:semantic-palette-guided-extraction}
In palette-based color editing, users modify the colors of an image by directly adjusting the corresponding palette. And each edited color $I'_i.C$ is defined as the original color $I_i.C$ plus a weighted sum of the palette colors' variations:
\begin{equation}
\label{eq:color-transfer}
    I'_i.C =  I_i.C + \sum^{|P|}_{j=1}w_j(I_i)(P'_j.C-P_j.C).
\end{equation}
Where $P_j$ and $P'_j$ are the palette entries before and after editing, $w_j(I_i)$ is the similarity weight of $I_i$ to $P_j$. Note that the palette only serves as an auxiliary medium in our pixel-level color editing. Specifically, we first find an optimal edited palette aligning with user edits, and then propagate local edits to similar regions with the solved semantic palette.


We define an energy function to measure the quality of the semantic palette $P'$ to be solved. It takes full account of the editing fidelity and the accuracy of color propagation. Therefore, our energy function is defined as the weighted sum of a fidelity term and a propagation term :
\begin{equation}
\label{eq:overall-loss}
    E = E_{\text{fidelity}} +  E_{\text{propagation}}. 
\end{equation}

The fidelity term forces user-stroke pixels to be accurately transferred to target colors. It is measured by the average L2 loss between the transferred and target colors:
\begin{equation}
\label{eq:fidelity}
    E_{\text{fidelity}} = \frac{1}{|H|}\sum_{i\in H} \| I'_i.C - \widehat{I_i}.C \|^2_2.
\end{equation}
Where $H$ is the index set of user-stroke pixels, $I'_i.C$ is the transferred color derived by the edited palette (Eq.\ref{eq:color-transfer}), and $\widehat{I_i}.C$ is the user-specified target color of $I_i$.

\begin{figure*}[htp]
\includegraphics[width=1\textwidth]{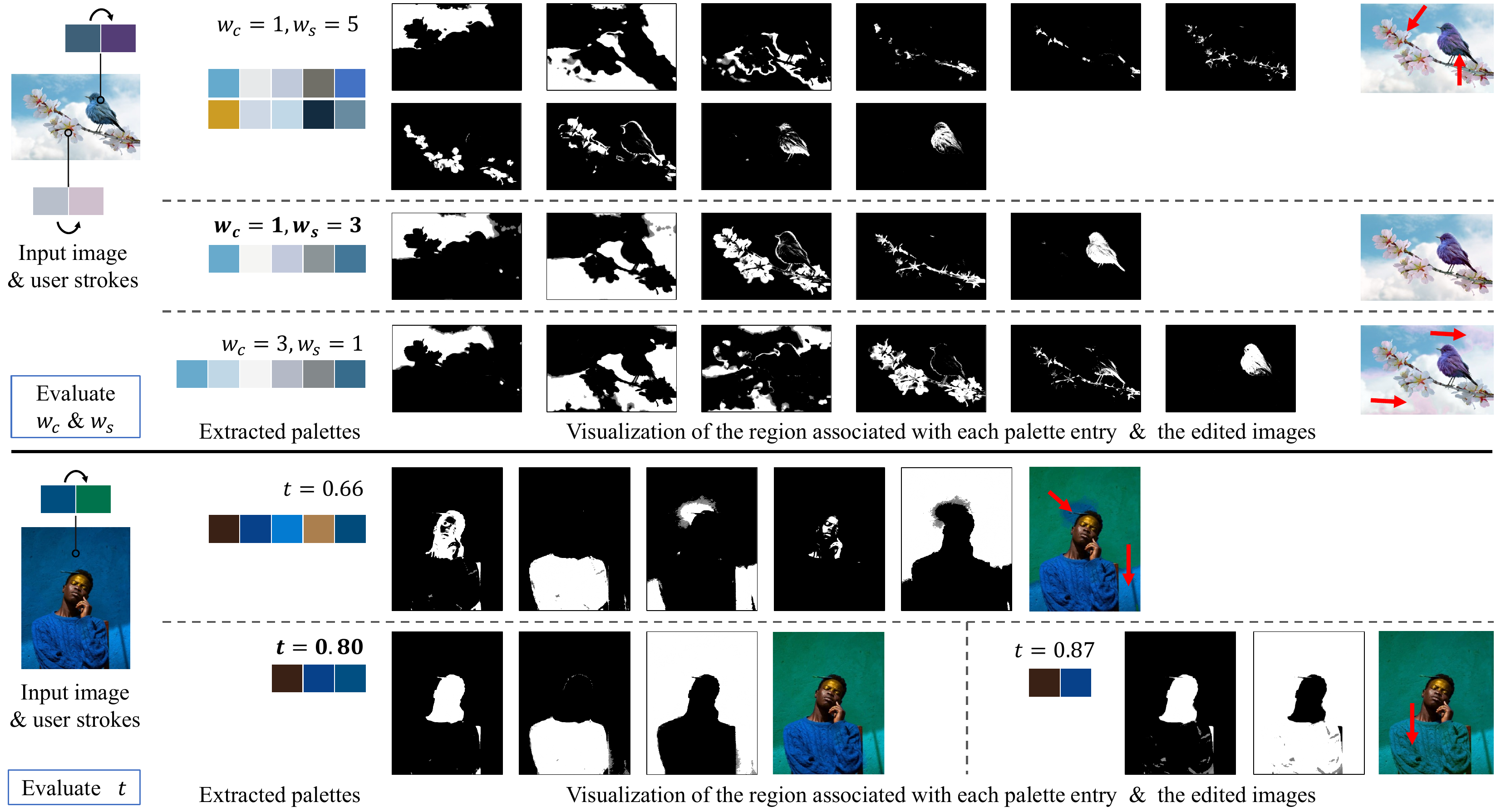}
\vspace{-0.3cm}
\caption{Parameter evaluation. The parameters $w_c$ and $w_s$ are evaluated in the first row, and $t$ is evaluated in the second row. For each example, we provide the input image and user strokes, the generated palettes with different parameter settings, region assigned to each palette entry, and the edited images.
}
\label{fig:param-eva}
\vspace{-0.5cm}
\end{figure*}

The propagation term ensures that user strokes are accurately propagated to semantically similar regions without affecting other parts. To this end, we design the following loss function to penalize those pixels have larger color variations but not similar with user strokes.
\begin{equation}
    E_{\text{propagation}} = \frac{1}{\sum_{i=1}^{|G|}\alpha_{j}}\sum_{j=1}^{|G|} \alpha_j \left\| I_j.C - I'_j.C \right\|_2^2.
\end{equation}
Where $G$ is the index set of sampled pixels. To accelerate calculation, we uniformly sample 256 pixels to form the sampling point set. $\alpha_j$ is a weight to measure the similarity between $I_j$ and user strokes, which is set as the similarity of $I_j$  to the most similar pixel in user strokes.



Once the edited semantic palette $P'$ is obtained, we employ the color transfer function (Eq.\ref{eq:color-transfer}) to map the palette colors’ variations to every pixel within the image in real time, resulting in content-aware color propagation.

\section{Experiments} 
\label{sec:experiments}
We performed all experiments on a laptop computer with AMD Ryzen 7 5800H 3.20 GHZ CPU and 16 GB RAM. We implement our approach in C++, and employ the NLopt library and the COBYLA solver to solve the edited palette.

\subsection{Parameter Evaluation}
In semantic palette extraction, we design a scheme to adaptively select the initial clustering centers for k-means. In which we use the sum of a color distance (weighted by $w_c$) and a semantic distance (weighted by $w_s$)  to measure the difference between a selected center and a candidate (Eq.\ref{eq:feature-distance}). Besides, we use a threshold $t$ to automatically stop the selecting process. 

For $w_c$ and $w_s$, combining a smaller $w_c$ and a larger $w_s$ usually yields more palette entries. However, due to semantic imprecision and lack of compensation for color information, the same object may be segmented to multiple detailed regions, and each region corresponds to a palette entry. Conversely, it generates fewer palette entries, but may result in different objects with similar colors sharing the same palette entry. As shown in the first row of  Fig.\ref{fig:param-eva}, when $w_c=1$ and $ w_s =5$, the extracted palette has 10 entries, but the cherry branch and the bird are assigned to multiple palette entries, thus user strokes on a petal fail to propagate to all petals. When $w_c=3$ and $ w_s =1$, the extracted palette has fewer entries, but some petals and the cloud share the same palette entry, it makes user strokes on the petals unexpectedly propagate to the cloud.

For the threshold $t$, a smaller value tends to produce palettes with more entries, but may cause the same region to be assigned to multiple palette entries. Conversely, a larger value typically generates a palette with fewer entries, however, different objects with similar colors may correspond to the same palette entry. As shown in the second row of Fig.\ref{fig:param-eva}, when $t=0.66$, it generates a palette with five entries, but the person is divided into three regions, and the wall is divided into two regions, causing user stroke on the top of the wall not to extend to the bottom of which. When $t=0.87$, it generates a palette with only two entries, but the sweater and the wall share the same palette entry, making user stroke on the wall incorrectly extends to the sweater.  

Overall, we find that setting the weight parameters $w_c=1$, $w_s=3$ and the threshold $t=0.80$ can generate palettes with an appropriate number of entries, and can lead to promising yet content-aware color propagation. So we use these values to extract semantic palettes for all examples. 



\subsection{Ablation study}
\begin{figure*}[htp]
\includegraphics[width=1\textwidth]{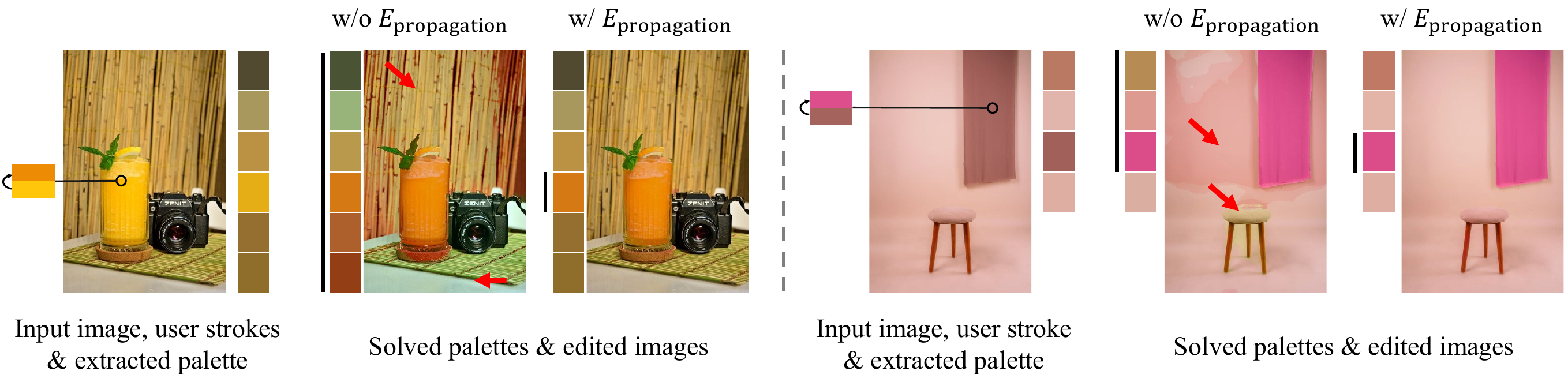}
\vspace{-0.6cm}
\caption{Ablation study. For each example, we provide the input image, user strokes, the generated palettes, the solved edited palettes, and the edited image.
}
\label{fig:ablation}
\vspace{-0.4cm}
\end{figure*}

In semantic palette-guided color propagation, we define an energy function (Eq.\ref{eq:overall-loss}) to measure the quality of the palette to be solved. It contains a fidelity term and a propagation term. The former is indispensable for interactive image editing.
While the latter ensures that user edits does not expand to regions of no interest, thus avoiding global color changes.

We perform an ablation study in Fig.\ref{fig:ablation} to validate the effectiveness of the propagation term. In the first example, the user wants to alert the color of the juice from yellow to orange. When removing the propagation term, the background and the floor appear to have an unnatural color change. In the second example, the user’s intention is to transform the fabric’s color from brown to pink. Yet, the removal of the propagation term results in an undesirable color modification of the wall and the bench. After adding the propagation term, it achieves more accurate color propagation without affecting other regions or objects.

From the ablation study, the propagation term plays a key role in the energy function, which can effectively avoid global color changes during color editing.

\subsection{Comparisons}
In this subsection, we qualitatively and quantitatively compare our method with five state-of-the-art ones. 

\begin{figure*}[htp]
\includegraphics[width=1\textwidth]{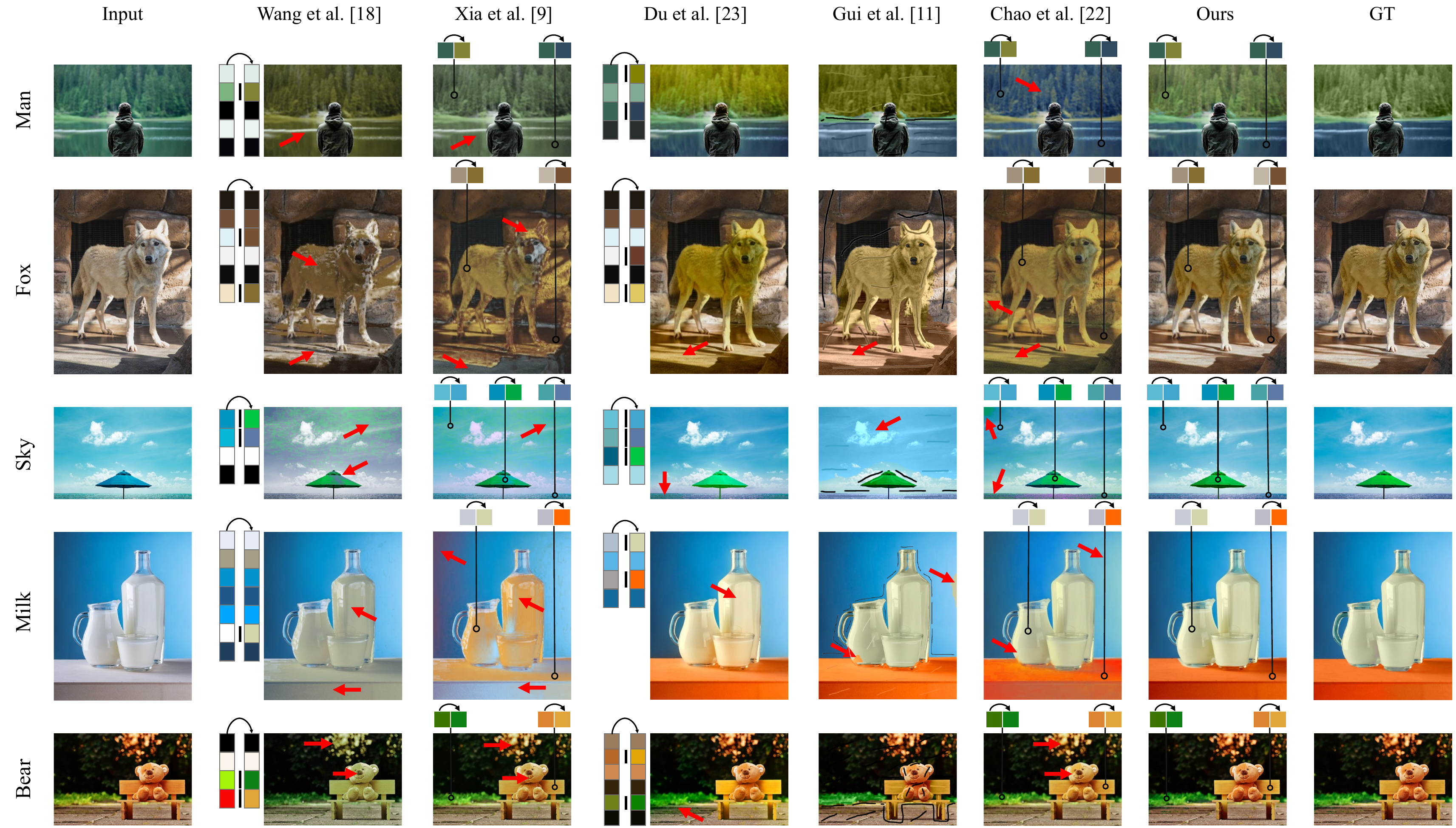}
\caption{Qualitative comparison of our method with other existing methods.
}
\label{fig:comparison}
\end{figure*}

The qualitatively comparison is shown Fig.\ref{fig:comparison}. Wang et al.\cite{WLX19} and Xia et al.\cite{Xia2024} computed an optimized convex hull in RGB space and used its vertices as color palettes. However, this approach does not take semantic information into account and thus inherently fails to achieve content-aware color propagation. As shown in the \textit{Man} sample, the user originally expected to alert the color of the forest to yellow, but the color of the lake also changed accordingly (row 1, columns 2-3).  While Du et al.\cite{du2024palette} integrated semantic features in color editing, their approach neither adaptively determines the number of palette entries nor supports pixel-level editing. So it's hard to avoid global color changes. As shown in the \textit{Milk} example, user stroke on the table unexpectedly expands to the bottle (row 4, column 4). Gui et al.\cite{GZ20} proposed the first deep learning-based color propagation. However, it struggles to capture similarities of distant pixels, it often requires user to provide dense strokes, and it is difficult to achieve good local control in color editing. As shown in the \textit{Sky} sample, user stroke on the umbrella are unnaturally extended to the cloud (row 3, column 5). Chao et al.\cite{chao2023loco} introduced a hierarchical semantic segmentation-based approach to color propagation, but this method heavily relies on the accuracy of segmentation, inaccurate segmentation will cause user strokes to propagate to unrelated regions. As shown in the \textit{Fox} example, inevitably, user stroke on the fox spreads to the stone (row 2, column 6). In contrast, our method accurately propagates user strokes to semantically similar regions.

The quantitative comparison is shown in Table \ref{table:comparison}. Compared with existing methods, the image generated by our approach achieves lower MSE,  
higher PSNR 
and higher SSIM.

\begin{table*}[]
\caption{Quantitative comparison of our method with other existing methods.}
\vspace{-0.2cm}
\resizebox{\textwidth}{10mm}{
\centering
\begin{tabular}{l|llllll|llllll|llllll}
\hline
\multirow{2}{*}{Example}  & \multicolumn{6}{c|}{MSE↓}   & \multicolumn{6}{c|}{PSNR(dB)↑}    & \multicolumn{6}{c}{SSIM↑}  \\ \cline{2-19}
 & Wang  & Xia  & Du   & Gui  & Chao & Ours     & Wang   & Xia     & Du    & Gui     & Chao  & Ours   & Wang  & Xia    &Du    & Gui  & Chao & Ours     \\
\hline
Man    	& 0.019 & 0.008 & 0.010 & 0.012 & 0.015 & \textbf{0.001} & 17.306 & 21.034 & 20.077 & 19.247 & 18.222 & \textbf{28.392} & 0.820 & 0.855 & 0.917 & 0.839 & 0.861 & \textbf{0.951}\\
Fox    	& 0.014 & 0.025 & 0.008 & 0.006 & 0.014 & \textbf{0.002} & 18.432 & 15.969 & 20.847 & 22.290 & 18.579 & \textbf{27.059} & 0.651 & 0.800 & 0.603 & 0.802 & 0.779 & \textbf{0.864}\\
Sky 	& 0.022 & 0.006 & 0.004 & 0.007 & 0.003 & \textbf{0.002} & 16.522 & 22.532 & 24.418 & 21.521 & 24.724 & \textbf{26.207} & 0.677 & \textbf{0.952}& 0.837 & 0.901 & 0.908 & 0.946 
\\
Milk   	& 0.027 & 0.043 & 0.004 & 0.003 & 0.004 & \textbf{0.002} & 15.729 & 13.652 & 23.843 & 25.150 & 23.681 & \textbf{27.231} & 0.846 & 0.922 & 0.722 & 0.925 & 0.861 & \textbf{0.934}\\
Bear    & 0.007 & 0.002 & 0.003 & 0.001 & 0.002 & \textbf{0.001} & 21.587 & 27.973 & 25.674 & 29.990 & 26.790 & \textbf{37.388} & 0.881 & 0.875 & 0.938 & 0.937 & 0.908 & \textbf{0.977}\\
\hline
\end{tabular}
}
\label{table:comparison}
\vspace{-0.4cm}
\end{table*}
\section{conclusion} 
\label{sec:conclusion}
This paper presents a novel semantic palette-guided approach to color propagation. To propagate local user edits to other regions with similar semantics, we first extract the semantic palette of an image from a high-dimensional feature space that integrates low-level color and high-level semantic information. We then  design an energy optimization method to find the edited palette aligning with user edits. At last, user edits are propagated to the whole image with the solved palette, achieving content-aware color propagation. 
Compared to existing color propagation methods, our approach enables efficient and accurate pixel-level color editing and ensures that local user edits are extended in a content-aware manner.

\section{acknowledgments} 
This work is supported by the Youth Program of the Natural Science Foundation of Qinghai Province (Project No. 2023-ZJ-951Q) and the National Natural Science Foundation of China (Project No. 62366043). 


\bibliographystyle{IEEEbib}
\bibliography{reference}

\end{document}